\documentclass[10pt,twocolumn,letterpaper]{article}

\usepackage[pagenumbers]{cvpr} 

\usepackage{graphicx}
\usepackage{amsmath}
\usepackage{amssymb}
\usepackage{booktabs}
 
\usepackage[pagebackref,breaklinks,colorlinks]{hyperref}

\usepackage[capitalize]{cleveref}
\crefname{section}{Sec.}{Secs.}
\Crefname{section}{Section}{Sections}
\Crefname{table}{Table}{Tables}
\crefname{table}{Tab.}{Tabs.}
\usepackage[colorlinks]{hyperref}

\def\cvprPaperID{*****} 
\def\confName{CVPR}
\def\confYear{2022}

\begin{document}
\hypersetup{urlcolor = [rgb]{0,0,0.5}, citecolor = [rgb]{0,0,0.5}}
\title{Snowpack Estimation in Key Mountainous Water Basins from Openly-Available, Multimodal Data Sources}

\author{Malachy Moran* \and Kayla Woputz* \and Derrick Hee* \and Manuela Girotto \and Paolo D'Odorico \and Ritwik Gupta \and Daniel Feldman \and Puya Vahabi \and Alberto Todeschini \and Colorado J Reed \\
}
\maketitle

%
\begin{abstract}
Accurately estimating the snowpack in key mountainous basins is critical for water resource managers to make decisions that impact local and global economies, wildlife, and public policy. Currently, this estimation requires multiple LiDAR-equipped plane flights or in situ measurements, both of which are expensive, sparse, and biased towards accessible regions.
In this paper, we demonstrate that fusing spatial and temporal information from multiple, openly-available satellite and weather data sources enables estimation of snowpack in key mountainous regions. Our multisource model outperforms single-source estimation by 5.0 inches RMSE, as well as outperforms sparse in situ measurements by 1.2 inches RMSE.
\end{abstract}

\section{Introduction}
\label{sec:intro}
Mountainous snowpacks are one of the most important water storage systems: directly providing an estimated 1.2 billion people with water for their agricultural, power, and personal consumption needs~\cite{barnett2005potential}.
In California, for instance, over 75\% of agricultural water supply comes from snow in the Sierra Nevada \cite{huning_climatology_2017}, which supports California's \$50.5 billion agriculture business that supplies 35\% of U.S.~vegetable production~\cite{pathak_climate_2018}. 
Similar reliance on snowpack can be found in many regions throughout the world: ranging from the snow-fed Colorado river, which drives an estimated \$1.4 trillion of economic activity~\cite{james2014economic}, to the Indus, Ganges, Brahmaputra, Yangtze, and Yellow river basins that support life and commerce in Asia~\cite{immerzeel2010climate}.

Accurately estimating the amount of snowpack in mountainous regions, as well as understanding how much of this snow will eventually become usable water, is a key, unsolved hydrologic problem motivating several large-scale, ongoing scientific studies~\cite{serreze1999characteristics,feldman2021surface}. 
While the underlying physics governing precipitation in mountainous regions is well understood, physically modeling this process in these regions is difficult due to their high variability in topography, surface characteristics, and complex basin and sub-basin formations which lead to highly variable physical processes and dramatically different snowpack~\cite{dettinger_winter_2004, huning_investigating_2018, lundquist_relationships_2010}. 
As a result, many hydrologists believe that making progress on this problem will require leveraging multimodal and multisource remote sensing data and machine learning~\cite{durand2021achieving}. 

The most accurate source of snowpack measurements come from in situ measurement devices, such as the network of 889 snow telemetry devices (SNOTELs)~\cite{serreze1999characteristics}. However, in situ measurements are often expensive (SNOTEL sensors cost $\$25-\$35$k and $\$3$k per year to maintain~\cite{patterson_2019}) 
and limited to regions that are human-accessible, preventing deployment at many important areas such as high-altitude regions that determine the expected water quantity late in the snow melt season~\cite{huning_investigating_2018,kirchner_lidar_2014}. 
In recent years, the Airborne Snow Observatory (ASO) flight missions have provided more complete geographic coverage in complex mountain systems by flying LiDAR-equipped plane flights over the regions. 
While these flights provide a wealth of high-resolution data, they are also expensive (e.g.,~one agency paid \$145k for two such flights and modeling~\cite{gazette_aso}), intermittent, and geographically limited to the regions accessible by the ASOs. 



In this paper, we fuse multiple, openly-available, daily satellite and weather data sources in order to predict high-resolution snowpack in key mountainous regions. Specifically, we use independent CNNs and LSTM encoders on data from the publicly available MODIS, Copernicus, Sentinel 1, and Sentinel 2 satellites as well as gridMET weather data and then fuse the encoded representations in order to predict Snow Water Equivalent (SWE) from ASO flights in water basins in the Sierra Nevada mountain range. Our multisource model outperforms single-source estimation by 5.0 inches RMSE, as well a outperforms sparse in situ measurements by 1.2 inches RMSE.



\begin{figure*}[t]
\centering
\includegraphics[width=.7\textwidth]{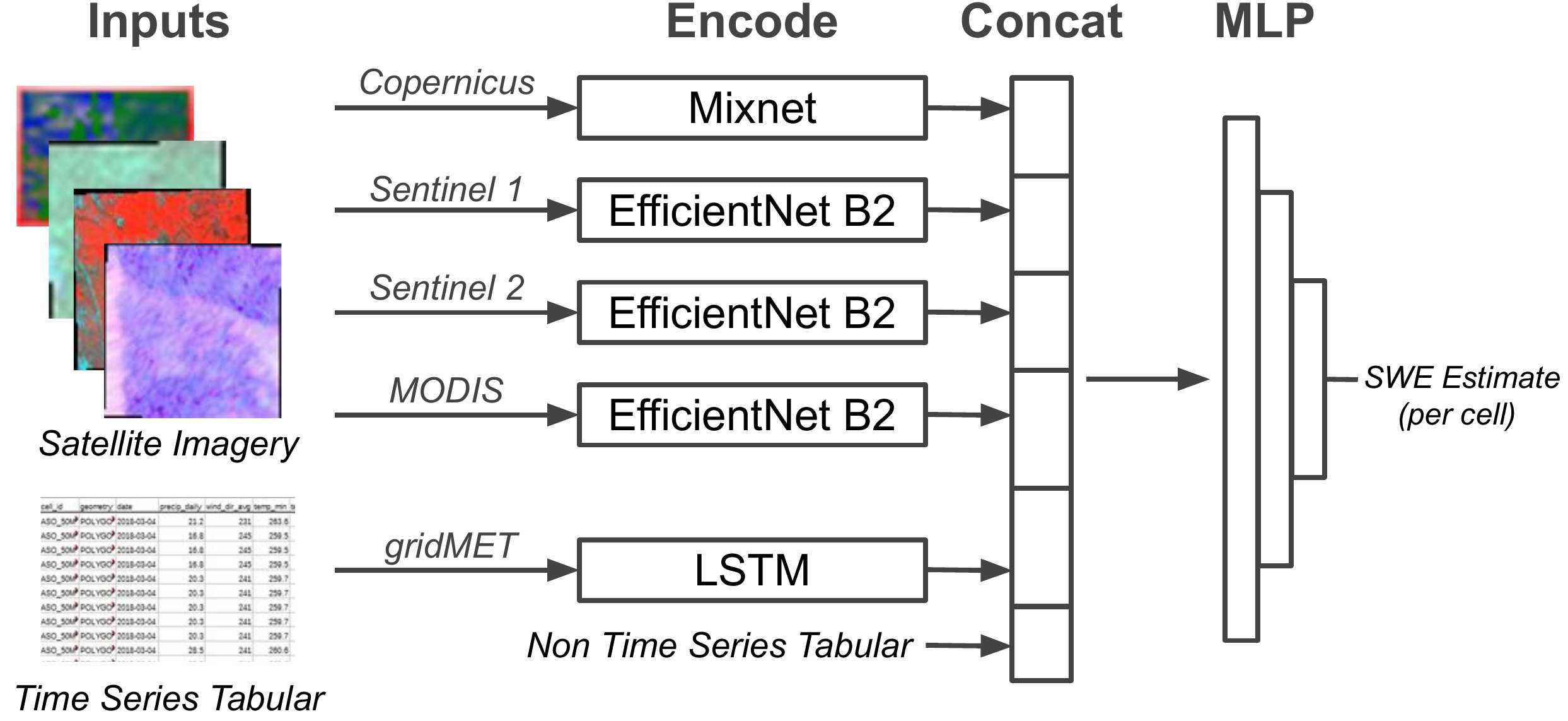}
\caption{Multimodal model for SWE estimation: Our model takes five sources of data, four are satellite images (Copernicus, Sentinel 1, Sentinel 2, MODIS), and one source is tabular weather data. The imagery are fed into convolutional neural networks, where Copernicus is fed into a smaller Mixnet model due to its simplicity, and Sentinel 1, Sentinel 2, and MODIS are fed into larger EfficientNet\_B2s. Time series gridMET data are fed into an LSTM, while the non-time-series are left unencoded. The output representations are then concatenated into a single vector which is fed through a 3-layer MLP in order to directly predict the SWE of each 1km$^2$ cell.}
\label{architecture}
\end{figure*}

\section{Background and Related Work}
\label{background-previous-work}
Mountainous regions represent a complex system with highly varied topography and surface characteristics.
Efforts to understand and investigate the physical models of various precipitation processes in these regions have a long history in hydrology, see e.g.,~\cite{huning_investigating_2018, dettinger_winter_2004, lundquist_relationships_2010, pulliainen1999hut}. 
Physical models assimilate observational data, such as in situ sensor measurements, which are often sparse both spatially and temporally, and at times inaccurate or unavailable. 
Furthermore, many models implicitly assume that point-scale in situ measurements are representative of large geographic regions. 
This assumption quickly degrades in complex topographic environments, such as mountainous regions, leading to inaccurate data assimilation and snowpack estimates~\cite{cayan_climate_2008}. 

One of the most accurate and geographically complete direct measurements of Snow Water Equivalent (SWE) comes from the Airborne Snow Observatory (ASO).
ASO was launched by NASA in 2013 to monitor SWE in key mountainous
basins in California and Colorado~\cite{behrangi_using_2018}. 
ASO flys a LiDAR-equipped airplane over these basins in order 
to measure surface elevation during snow-free periods as a
baseline, and then again during the snow periods, yielding snow depth as
the difference. ASO converts this snow depth into SWE by using snow density values derived from a snow
model, supported by available ground measurements~\cite{behrangi_using_2018}. 
While the ASO flights have excellent coverage of geographical variability, they are expensive and infrequent. 

Several works have explored hybrid machine learning and physical modeling techniques to estimate and predict snowpack in mountainous regions from openly available satellite data~\cite{meyal_automated_2020,duan_comprehensive_2021,bair2018using}. 
Specifically, a number of works addressed the temporal and spatial sparsity by incorporating Landsat images~\cite{williams2006landsat} to measure fractional snow-covered
area and then feed these measurements into a Bayesian data assimilation framework to gather
estimates of SWE on a daily 90m scale~\cite{huning_climatology_2017,margulis_particle_2015, margulis_landsat-era_2016}. 
These works showed that accurate SWE estimates must account for high spatial variability, that solely relying on in situ measurements biases towards lower elevations, and that wind speed and moisture are key variables for estimating snowpack in these regions. 

In \cite{duan_comprehensive_2021}, the authors explored a purely machine learning approach to estimating SWE using three techniques: LSTM, Temporal CNN, and attention-based models. They applied each model to a time series of SWE measurements from individual SNOTEL sites and then extrapolated the resulting predictions to a 4km square spatial resolution grid at a daily temporal resolution. Their work showed that deep learning models can obtain quality estimates of SWE prediction. However, their work had similar constraints as
earlier models by not being able to account for the geographic
variability desired. In addition, their use of SNOTEL sites opens the estimates and predictions up to the biased issues with SNOTEL data regarding low elevation and inability to capture a representative sample
of dynamic geographies.


The United States Bureau of
Reclamation hosted a ``Snowcast Showdown'' data science competition in order to predict SWE at various in-situ measurement locations.\footnote{\url{https://www.drivendata.org/competitions/86/competition-reclamation-snow-water-dev/}} This competition began November 2021 and concludes
June 2022.  With a large prize purse of \$500k, this shows the growing
interest and willingness for investment in tackling this critical
problem space.


\section{Data and Methodology}
Deep learning has been applied to a wide variety of environmental prediction and estimation problems in recent years, see \cite{yuan_deep_2020}. Here we focus on the following prediction problem: given a set of openly available, daily data sources, we aim to predict accurate estimate of Snow Water Equivalent (SWE) in key mountainous basins. 

Table~\ref{data-source} summarizes the daily, openly available data sources we use as input: satellite data from the MODIS, Copernicus, Sentinnel 1 and 2, and weather data from gridMET. We target SWE measurements from the Airborne Snow Observatory (ASO) as the ground truth, where we focus our study on nine basins in the Sierra Nevada range. Selection of these basins was driven by the data available through the National Snow and Ice Data Center (NSIDC) Airborne Snow Observatory (ASO) dataset (see appendix). These basins represent approximately 28,750 km$^2$ of the Sierra Nevada Region.

\begin{table}[h]
\includegraphics[width=0.48\textwidth]{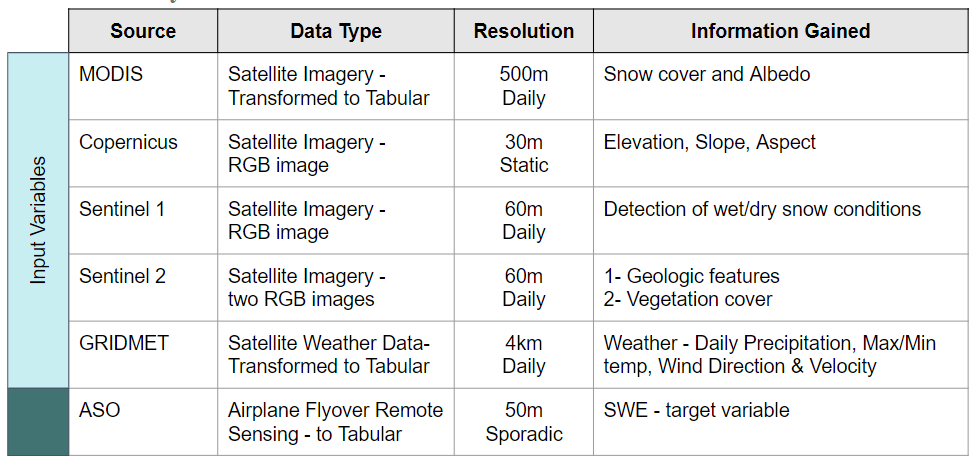}
\caption{Summary of data sources, data types, resolution, and variables used from each. For a deeper discussion on data sources, manipulations and comments please see Appendix A.}
\label{data-source}
\end{table}

\begin{table}[t]
\centering
\begin{tabular}{ |p{2.2cm}||p{1.1cm}|p{1.1cm}|p{1.4cm}|  }
Basins & Basin Size $10^3km^2$ & SWE Mean $\pm$ Std (\textit{in.}) & RMSE (\textit{in.}) \\
\hline
 Feather & 8.4 & 1.2$\pm$3.7 & 2.7 \\
 Yuba & 2.2 & 5.8$\pm$8.8 & 7.0 \\
 Truckee & 2.9 & 7.4$\pm$8.8& 9.4 \\
 Carson & 1.5 & 6.8$\pm$8.5 & 7.3 \\
 Tuolumne & 2.9 & 5.6$\pm$9.1 & 9.5 \\
 Merced & 1.7 & 7.2$\pm$7.6 & 5.8 \\
 San Joaquin & 4.2 & 6.3$\pm$8.7 & 7.6 \\
 Kings Canyon & 3.5 & 8.0$\pm$7.7 & 17.5 \\
 Kaweah & 1.5 & 3.1$\pm$7.7 & 5.1 \\
 \hline
 \textbf{Overall} & \textbf{28.8} & \textbf{4.0$\pm$7.0} & \textbf{7.5} \\
 \hline
 \hline
 Zero Pred & - & - & 8.7 \\
 Mean Pred & - & - & 13.0 \\
 SNOTEL & - & - & 8.7
\end{tabular}
\caption{Experimental Results: This table's columns shows basin name, area, ground truth SWE in the basin from ASO, and RMSE of our estimated SWE content. The bottom shows three baseline prediction techniques: predicting zero values (no snow), predicting the mean value of the training data, and predicting the value of the nearest SNOTEL station. Our multimodal fused model outperforms each of these baselines.}
\label{results}
\end{table}

We fuse the multimodal data sources in the following way as demonstrated in Figure~\ref{architecture}: each satellite image data source is encoded using an independent CNN encoder (we use smaller MixNet~\cite{tan2019mixconv} for the simpler, non-changing Copernicus data and EfficientNets~\cite{tan2019efficientnet} for the time-varying satellite images). 
Time series gridMET data are fed into an LSTM~\cite{hochreiter1997long}, while the non-time-series are left unencoded. The output representations are then concatenated into a single vector which is fed through a 3-layer MLP in order to directly predict the SWE of each 1km$^2$ cell.

For this initial investigation, we predict at a computationally-convenient 1km$^2$ resolution compared to the 50m$^2$ resolution of the original ASO imagery, and look to explore higher resolutions outputs in future work. 
Furthermore, we found that boundaries between predicted cells were a source of high error. 
Therefore, we apply a post processing Gaussian smoothing to the predicted output, which smooths the borders between predicted cells, and led to lower prediction error in our early investigations.

The key variables fed into the CNNs include elevation, slope, and aspect (compass direction) of the surface, infrared measures to detect wet and dry snow conditions, geologic feature detection, and vegetation cover. Other variables are pulled in from satellite imagery, saved in tabular form, and fed into our LSTM, namely snow cover, albedo, daily total precipitation, maximum temperature, minimum temperature, wind direction, and wind velocity.

\section{Results}
Our model was trained on data spanning 2016-2019. To calculate an observed RMSE on an unseen test set, we use the most recent ASO flyovers of each basin, which span March-April of 2022.
As shown in Table~\ref{results}, we observe a mean test RMSE of 7.5 inches, where the mean has been weighted by the area of each basin. The RMSE observed significantly varies by basin, where for instance, the Feather basin has a mean RMSE of 2.7 inches, while the Kings Canyon basin has a mean RMSE of 17.5 inches. 

\begin{figure}[t]
\centering
\includegraphics[width=.35\textwidth]{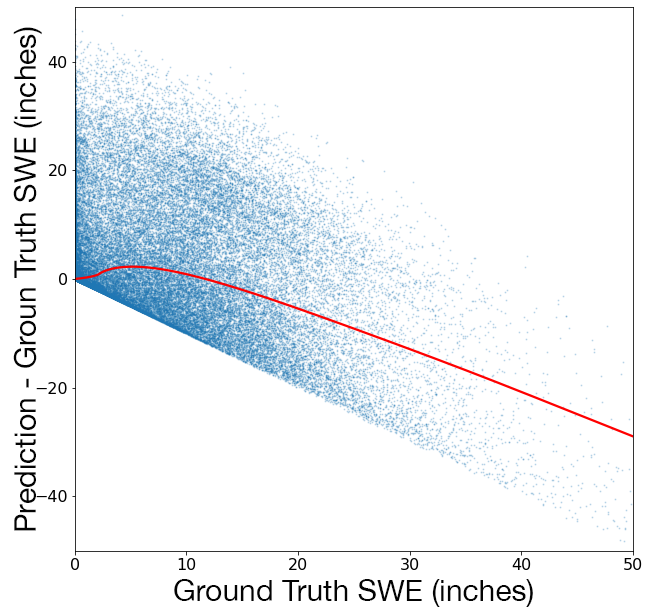}
\caption{SWE vs Predicted Error: The x-axis shows the ground truth SWE at each 50m$^2$ cell, the y-axis shows the predicted value minus the ground truth SWE for 100k randomly sampled locations, and the red line shows a Loess regression. This plot shows that our model tends to underpredict SWE for high ground truth SWE values and tends to overpredict SWE for low SWE values.}
\label{fig:pred-error}
\end{figure}

Figure~\ref{fig:pred-error} shows the actual SWE vs the prediction error and indicates the reason for this large performance difference between basins: our model tends to underpredict SWE for high ground truth SWE values and tends to overpredict SWE for low SWE values. As shown by the mean values in Table~\ref{results}, the Feather basin has a much lower average SWE value than Kings Canyon, where our model tends to slightly overpredict the SWE for Feather and largely overpredict the SWE for Kings Canyon. 

The bottom of Table~\ref{results} shows three baseline prediction techniques: predicting zero values (no snow), predicting the mean value of the training data, and predicting the value of the nearest SNOTEL station. Surprisingly, predicting the nearest SNOTEL station value tends to be as inaccurate for a given location as simply predicting no snow. In the case of the Feather basin, the small amount of snow in the test regions differ from the larger amount of snow in the SNOTEL measurement sites, while the large amount of snow in Kings Canyon is reflected by the SNOTEL measurement site. These offset overpredictions are inverted in the case of predicting no snow, i.e. the small amount of snow in the Feather region has low error while the high amount of snow in Kings Canyon has high error.


Table~\ref{ss-results} compares the multi-source model with single source models for each individual source, using a held-out validation set from 2016-2017. 
Our multisource model outperforms single-source estimation by 5.0 inches RMSE, i.e.,~each of the individual models have low performance -- RMSE values $> 14$ -- whereby combining all sources, the final RMSE is reduced to 9.1. This indicates that multisource information is necessary to jointly model and estimate SWE.

\begin{table}
\centering
\begin{tabular}{ |p{2.2cm}||p{1.1cm}| p{1.1cm}|  }
Source Data & model & RMSE  (\textit{in.}) \\
\hline
 Copernicus & Mixnet & 14.9 \\
 Sentinel 1 & EffNet & 14.1 \\
 Sentinel 2 & EffNet & 14.7 \\
 MODIS & EffNet & 15.1 \\
 gridMET & LSTM & 14.4 \\
 \hline
 All & Fusion & \textbf{9.1}
\end{tabular}
\caption{Multi vs single source: This table compares the multi-source model with single source models for each of the five input sources.
Each of the individual models obtain RMSE values $> 14$. Combining all sources via Figure~\ref{architecture} leads to an RMSE of 9.1, indicating that multisource information improves the ability to jointly model and estimate SWE in these regions.}
\label{ss-results}
\end{table}




\section{Future Work \& Considerations}
This initial study shows the promise of leveraging multimodal satellite and weather data to estimate SWE in key mountainous basins. We see the following as promising directions to pursue in order to both improve performance as well as scientific understanding of SWE estimation:

\begin{itemize}
    \item Improved resolution - predicting SWE at high resolutions, such as 50m resolution, can better leverage the resolution of the input data as well as the overlapping regions between multiresolution input sources.
    \item Modify our snow albedo inputs. There is a rich literature around snow albedo, where a great deal of work has been done to create corrections between this value and snow estimation. We do not currently incorporate these understandings in our modeling.
    \item Incorporate the GEDI satellite imagery to get more accurate vegetation coverage. GEDI utilized LiDAR to measure canopy height, ground cover, and more. This is a highly more accurate data source than our current Sentinel-2 vegetation bands. 
    \item Reevaluation of wind metrics. Currently our wind and aspect values are run in different parts of our model. Its been shown that  wind direction in the Sierras is a better predictor of snow accumulation than precipitation, however it is the angle of intersection with the ground that is important. With our wind direction and aspect pulled out separately, this important feature is not factored into our model.
\end{itemize}

 We additionally plan to explore more multimodal and multiscale data sources as well as modeling approaches to increase our temporal coverage and better capture the unique seasonal impacts between years.

\section{Conclusion}
Here we present a prediction tool that uses free and available data sources to estimate snow water equivalent (SWE) in the Sierra Nevadas within an RMSE of 7.5” when compared to ASO measurements. Being able to accurately estimate the SWE is critical for water management, policy makers, and researchers. This need becomes more and more critical as the climate-related water crises take place around the world. We have packaged our analysis code and data loaders into a convenient (``pip-installable'') python package available at \url{https://github.com/Seiris21/ucb2022.snowcast}.

{\small
\bibliographystyle{ieee_fullname}
\bibliography{egbib}
}

\appendix


\section{Data Source Details}

\subsection*{SNOTEL}

\textbf{Data Source Location:} \url{https://www.nrcs.usda.gov/wps/portal/wcc/home/snowClimateMonitoring/snowpack/}  

\textbf{Description of Data Source:} SNOTEL is a set of remote monitoring stations managed by the USDA Natural Resources Conservation Service National Water and Climate Center. They are physical stations installed on site throughout 12 states on the West Coast of the U.S.. The sites measure SWE, Snow Depth, and the change in each.  

The measurements from SNOTEL are available as .csv files as either daily, weekly, or monthly data.

\textbf{Variables/Data Used:} No data from SNOTEL was used in the final model.

\textbf{Transformations/Modifications:} No data from SNOTEL was used in the final model.

\textbf{Discussion:} When this project was originally formulated, SNOTEL data was incorporated as a possible input or target variable. This proved to be problematic for several reasons.

Firstly, there is an issue of spatial coverage. Because SNOTEL stations are physically installed on-site, they do not represent an unbiased sample. A human being must be able to reach the site in question to install the station, and the location must be capable of supporting a station. This means that SNOTEL stations aren’t often located in places of particular interest, such as near the peaks of mountains. There is also an issue of labor. Unlike a plane or a satellite, which can cover large swaths of land in a single image or sweep, each SNOTEL station must be installed individually. As such they are often spatially sparse, and measure only in their immediate vicinity. This creates large spatial holes in the area that the available data covers.

Secondly, there is an issue of temporal coverage. Exploratory data analysis showed that there were large gaps in reporting from stations. This makes SNOTEL problematic as input variable for the model, because it may mean that on the date that is being predicted the nearest SNOTEL station that is actively reporting could be more than a hundred miles away. It also makes SNOTEL difficult to use as a target variable, as we cannot judge whether the reason a station is not reporting is random, or caused by a factor that may be significant to SWE, such as a storm.

\textbf{Alternative Data Sources:} Data collected from the Airborne Snow Observatory was used instead of SNOTEL data.

\subsection*{Airborne Snow Observatory (ASO)}

\textbf{Data Source Location:}

\url{https://data.airbornesnowobservatories.com/}

\url{https://nsidc.org/data/aso}

\textbf{Description of Data Source:} The Airborne Snow Observatory, first managed by the National Snow and Ice Data Center (NSIDC) and now a private enterprise, conducts airborne remote sensing campaigns over the western United States. Using airplane mounted lidar and hyperspectral sensor equipment, ASO measures variables on a sporadic basis during the months of expected peak snowpack. Variables it measures include Snow Depth at the 3m and 50m resolution, as well as Snow Water Equivalent at a 50m resolution. Variable measurements are targeted at “basins” or particular geographic areas that drain into specific watersheds. Variable values are collected in meters. Areas in the image not in the basin are coded as null or NA values.

The measurements from ASO are available as .tiff images, where the pixel values represent the variable in question. The images also contain latitude and longitude data about each pixel. The campaign was conducted on an annual basis from 2013-2019 by the NSIDC and from 2020 to the present by a private company.

\textbf{Variables/Data Used:} The Snow Water Equivalent measurements at the 50m resolution were consolidated into 1km blocks (see Transformations/Modifications) and used as the target variable for training. The latitude and longitude from these blocks was also collected to define the boundaries of other satellite imagery. Due to the available time windows of other data sources, only the years of 2016-2019 were used as training data for the model. 2020-2022 were reserved as an unseen test set.

\textbf{Transformations/Modifications:} The Snow Water Equivalent measurements were transformed from meters to inches in order to comport with other training data sources. Then the data was converted from a 50m to a 1km resolution by taking the average SWE value of a 20 pixel by 20 pixel block.

\textbf{Discussion:} There are many possible sources from which to obtain Snow Water Equivalent for the purposes of creating a training target variable or estimating total water available in the snowpack. Multiple satellites attempt to infer this data from infrared and visible light measurements of the snowpack. However, according to the advising professors on this project, the data collected from ASO is the most accurate. This is due to the fact that it is conducted via airplane flyover with specialized equipment, rather than via satellite. 

Despite this increased accuracy, ASO has some major limitations. Its primary advantage, the fact that it is a flyover mission, is also its achilles heel. Bad weather or dangerous flight conditions can delay data collection. It is also a relatively expensive process, as multiple flights may be needed to survey one basin, where a satellite can image the entire range in a single frame. For this reason flights are only conducted a few times a year for each basin, sometimes only once.

It is these limitations that make ASO amenable to replacement with a Machine Learning Framework. If the data already collected can be used to train an AI model, this could reduce or eliminate the need for costly flyovers, while maintaining the high level of accuracy.

\textbf{Alternative Data Sources:} As mentioned in the discussion, SNOTEL, CDEC, and HRRR, along with many other sources have estimated Snow Water Equivalent and Snow Depth measurements.

\textbf{Reference:} 

Painter, T. 2018. ASO L4 Lidar Snow Water Equivalent 50m UTM Grid, Version 1. [Indicate subset used]. Boulder, Colorado USA. NASA National Snow and Ice Data Center Distributed Active Archive Center. https://doi.org/10.5067/M4TUH28NHL4Z.

\subsection*{MODIS Terra/Aqua Snow Cover Daily Global 500m}

\textbf{Data Source Location:}

\url{https://developers.google.com/earth-engine/datasets/catalog/MODIS_006_MOD10A1}

\url{https://developers.google.com/earth-engine/datasets/catalog/MODIS_006_MYD10A1}

\url{https://modis.gsfc.nasa.gov/about/specifications.php}

\textbf{Description of Data Source:} The MODIS (Moderate Resolution Imaging Spectroradiometer) instrument aboard the NASA Terra and Aqua satellites collects information about albedo and snowcover in the 459-2155nm wavelength range, imagining the entire planet approximately every 1-2 days at a 500m resolution. Data is provided in the form of snow cover estimates, snow albedo, fractional snow cover estimates, and image quality data. Snow cover data is estimated using a Normalized Difference Snow Index (NDSI) test.

\textbf{Variables/Data Used:} The albedo, snow cover, and raw NDSI variables from both MODIS satellites are used. Quality data was not used.

\textbf{Transformations/Modifications:} Data was captured for each 1km square and normalized to be between 0 and 1 dividing the raw value by its theoretical maximum. Because of the low resolution of the MODIS data (500m), only four pixels would be observed for each 1km square. Therefore the mean of the four pixels was taken, transforming the data into a tabular format.

\textbf{Discussion:} When MODIS was originally included in the model, an actual image from each of the satellites was used. The image was formatted as an RGB image, with each color channel representing one of the three variables taken from the satellite. Due to MODIS’s low resolution, this was discarded. In order to get even a small image, multiple kilometers around the area of interest also had to be brought into the model. This posed a particular problem when it came to managing GPU memory, as using MODIS imagery required the creation of another image model. It was decided that it wasn’t worth taking up valuable memory space processing large areas outside of our area of interest when that space could be used to fit larger and more powerful image models.

Instead, we limited MODIS observations to only the square kilometer in question, and transformed the data into tabular format. This allows the MODIS data to be added after the image models and the LSTM had been combined into a linear layer.

\textbf{Alternative Data Sources:} There are multiple other sources that could be used to capture Albedo or surface reflectance. \href{https://planetarycomputer.microsoft.com/dataset/landsat-8-c2-l2}{Landsat 8} captures surface reflectance measures, as does the \href{https://developers.google.com/earth-engine/datasets/catalog/COPERNICUS_S3_OLCI}{Ocean and Land Color Instrument (OLCI)} on the Sentinel 3 satellite. Many of these products suffer from the same issues of low resolution as MODIS.

\subsection*{Copernicus DEM GLO 30 - Digital Surface Model}

\textbf{Data Source Location:}

\url{https://planetarycomputer.microsoft.com/dataset/cop-dem-glo-30}

\textbf{Description of Data Source:} The Copernicus Digital Elevation Model/Digital Surface model is a representation of the Earth’s surface, including man made features and to some extent vegetation, created by recording its elevation. The data for the model comes from a radar instrument on the TanDEM-X Mission, which was funded jointly by the German government and Airbus. The images from Copernicus are static, and do not change over time.

\textbf{Variables/Data Used:} There is only one variable that is measured by Copernicus, and that is elevation. It has one of the most detailed resolutions of any of our data sources, being available down to 30 meters. Elevation data was extracted from Copernicus for each 1km square once for the entire training period. This data was captured as a 1 channel image, with the value of each pixel being the measured elevation.

\textbf{Transformations/Modifications:} Using the elevation measurements, we were able to construct two additional variables. The slope and aspect of each pixel was calculated using the datashader python package, based on the elevation values of the pixels surrounding it. These two variables were added as two additional channels to create an RGB image.

\textbf{Discussion:} When first constructing the model, the Copernicus 90m resolution DEM was used. After some initial model training and a literature review, it was decided to use the 30m model instead, as it appeared that elevation, slope and aspect would be significant for snow water equivalent.

Our primary concern with Copernicus is that given our model structure, we currently do not incorporate both wind direction and aspect simultaneously in the model. Multiple sources (2018, Huning et al. 2017, Margulis et al. 2016, Margulis et al. 2015) suggest that these two variables matter most when taken into consideration together.

\textbf{Alternative Data Sources:} Copernicus is also available at the 90m resolution, and there are many alternative DEMS available. This includes the \href{https://planetarycomputer.microsoft.com/dataset/3dep-seamless}{USGS 3DEP Seamless DEM} which can be obtained at a 1m resolution for some areas.

\subsection*{Sentinel-1 SAR GRD: C-band Synthetic Aperture Radar Ground Range Detected, log scaling}

\textbf{Data Source Location:}

\url{https://developers.google.com/earth-engine/datasets/catalog/COPERNICUS_S1_GRD}

\textbf{Description of Data Source:} Sentinel-1 is operated by the European Space Agency (ESA) and was introduced as a radar vision component of the Copernicus program. It is equipped with two polar orbiting satellites, which perform a synthetic aperture with radar imaging (SAR). Sentinel-1 bands can collect data in all weather conditions.

\textbf{Variables/Data Used:} Three bands from the Sentinel-1 data are pulled to form a single RGB image for our modeling purposes: VV, VH, and VV+VH. The combination of these bands are commonly used to detect wet/dry snow conditions. Data was pulled at the 40 meter resolution. 

\textbf{Transformations/Modifications:} No transformations were done by our team, however Google Earth Engine conducts pre-processing including the following steps: thermal noise reduction, radiometric calibration, and terrain correction using SRTM 30 or ASTER DEM where final terrain-corrected values are converted to decibels via log scaling.

\textbf{Discussion:} Sentinel-1 SAR data is used to determine the presence of wet/dry snow conditions, crop health, and more. While there is no direct alternative or preferred datasource for collecting this data, we do still plan to undergo additional experimentation and testing to evaluate the impact this datasource has on our modeling. 

\textbf{Terms of Use:} The use of Sentinel data is governed by the \href{https://scihub.copernicus.eu/twiki/pub/SciHubWebPortal/TermsConditions/Sentinel_Data_Terms_and_Conditions.pdf}{Copernicus Sentinel Data Terms and Conditions}.

\subsection*{Sentinel-2 MSI: MultiSpectral Instrument, Level-1C}
\textbf{Data Source Location: }

Sentinel-2 MSI: MultiSpectral Instrument, Level-1C  |  Earth Engine Data Catalog  |  Google Developers

\textbf{Description of Data Source:} Sentinel-2 is a wide-swath, high-resolution, multi-spectral imaging mission supporting Copernicus Land Monitoring studies, including the monitoring of vegetation, soil and water cover, as well as observation of inland waterways and coastal areas. This satellite is operated by the European Space Agency (ESA) as a part of the Copernicus mission.

\textbf{Variables/Data Used:} Sentinel-2 has a wide variety of bands that with the right combinations can be used to visualize distinct aspects of the survey area. For this project we created two sets of RGB images that visualize the geologic features and vegetation for the study area. The geologic features were created by assigning our three image channels to bands B8A (Red Edge4), B12 (SWIR 2), and the ratio of the two as calculated by (B8A-B12)/(B8A+B12). The vegetation images were created by assigning our three image channels to bands B2 (Blue), B4 (Red), and a calculated band incorporating B8 (NIR), calculated as (B8- B2)/(B8+B2). Images were pulled at the 20m and 60m resolutions.

\textbf{Transformations/Modifications:} No transformations were done on the data set with the exception of creating the calculated bands noted above. The data set does have a cloud masking tool provided through Google Earth Engine which allows for flexible modification of cloud sensitivity.

\textbf{Discussion:} A great deal of optimization still remains for this dataset. When creating the calculated bands, the resulting RGB images are heavily skewed towards two color channels. This skewness is likely minimizing the impact these images have on the model. Further research into the proper scaling of each channel to optimize the signal needs to be completed for our next iteration of modeling. At minimum, normalizing values should be completed to provide a more balanced output.

\textbf{Alternative Data Sources:} We’re considering an alternate datasource to capture more accurate vegetation coverage, as well as canopy height information. Switching the vegetation bands to the \href{https://gedi.umd.edu/}{GEDI LiDAR} source could greatly improve our vegetation input, however wrangling this data source could prove computationally heavy. On our next iteration we will be working to bring in this data and run experiments to identify its relative importance in the model as well as efficiency for the prediction tool.

\textbf{Terms of Use:} The use of Sentinel data is governed by the \href{https://scihub.copernicus.eu/twiki/pub/SciHubWebPortal/TermsConditions/Sentinel_Data_Terms_and_Conditions.pdf}{Copernicus Sentinel Data Terms and Conditions}.

\subsection*{gridMET - University of Idaho Gridded Surface Meteorological Dataset}
\textbf{Data Source Location:}

\url{https://developers.google.com/earth-engine/datasets/catalog/IDAHO_EPSCOR_gridMET}

\textbf{Description of Data Source:} The gridMET Gridded Surface Meteorological Dataset is a relatively high (for weather data) 4km resolution dataset. It blends together information from both the PRISM tool on the ALOS satellite with data from the National Land Data Assimilation System (NLDAS) which is maintained by NASA to produce a spatially and temporally continuous mapping of weather across the continental United States. The dataset is updated with provisional measurements as they become available, these are replaced with the final measurements after a few days.

\textbf{Variables/Data Used:} The variables used from this dataset include total precipitation, maximum and minimum temperature, average wind direction and average wind velocity, for each day. The data was captured in tabular format for the date of the target SWE measurement from ASO, as well as the 10 preceding days. This entire table of 10 days worth of data was used to feed into the LSTM model.

\textbf{Transformations/Modifications:} No transformations were done on the data set.

\textbf{Discussion:} Meyal, A. Y. et al (2020) suggests that weather data fed into an LSTM on its own is a powerful SWE prediction tool. As such, obtaining weather data was a primary concern for modeling. The need for full geographic and temporal coverage of specific weather measurements from an easily accessible datasource pushed us to gridMET. This does not necessarily mean it is the most accurate or best source. It has a relatively limited set of 16 possible weather measurements to choose from, where other sources have over 100. That being said, it is accessible through Google Earth Engine with minimal transformation and a high rate of availability. Exploration of alternative resources to gridMET could improve the model.

\textbf{Alternative Data Sources:} There are several alternative available resources to the gridMET data source. One that we strongly considered was the \href{https://rapidrefresh.noaa.gov/hrrr/}{High Resolution Rapid Refresh (HRRR)} dataset, which is maintained by the National Oceanic and Atmospheric Administration (NOAA). This source has a slightly better resolution at 3km, offers an overwhelming number of variables, and has the advantage of incorporating radar data from NOAA’s Rapid Refresh radar.

\end{document}